\documentclass[10pt,twocolumn,letterpaper]{article}

\usepackage{cvpr}
\usepackage{times}
\usepackage{epsfig}
\usepackage{graphicx}
\usepackage{amsmath}
\usepackage{amssymb}

\usepackage{lipsum}
\usepackage[]{booktabs}
\newcommand{\smalltitle}[1]{\vspace{0.2em}\noindent \textbf{{#1}}}

\newcommand{\I}{\mathbf{I}}

\newcommand{\bS}{\mathbf{S}}

\newcommand{\bz}{\mathbf{z}}

\newenvironment{myitemize}[1][]{
	\begin{list}{{#1}} 
		{
			\setlength{\leftmargin}{1.2em}
			\setlength{\topsep}{0em}
			\setlength{\itemsep}{-0.2em}
	}}
{\end{list}}

\usepackage[pagebackref=true,breaklinks=true,letterpaper=true,colorlinks,bookmarks=false]{hyperref}

\cvprfinalcopy 

\def\cvprPaperID{1343} 

\ifcvprfinal\fi
\begin{document}

\title{3D Human Pose Estimation in the Wild by Adversarial Learning}

\author{
Wei Yang$^{1}$ \quad 
Wanli Ouyang$^{2}$ \quad 
Xiaolong Wang$^{3}$ \quad 
Jimmy Ren$^{4}$ \quad 
Hongsheng Li$^{1}$ \quad
Xiaogang Wang$^{1}$ \\ \\
$^{1}$ CUHK-SenseTime Joint Lab, The Chinese University of Hong Kong\\
$^{2}$~ School of Electrical and Information Engineering, The University of Sydney \\
$^{3}$~ The Robotics Institute, Carnegie Mellon University \\
$^{4}$~ SenseTime Research 
}

\maketitle
\thispagestyle{empty}

\begin{abstract}
	Recently, remarkable advances  have been achieved in 3D human pose estimation from monocular images because of the powerful Deep Convolutional Neural Networks (DCNNs). 
	Despite their success on large-scale datasets collected in the constrained lab environment, it is difficult to obtain the 3D pose annotations for in-the-wild images. 
	Therefore, 3D human pose estimation in the wild is still a challenge. 	
	In this paper, we propose an adversarial learning framework, which distills the 3D human pose structures learned from the fully annotated dataset to in-the-wild images with only 2D pose annotations. 
	Instead of defining hard-coded rules to constrain the pose estimation results, we design a novel multi-source discriminator to distinguish the predicted 3D poses from the ground-truth, which helps to enforce the pose estimator to generate anthropometrically valid poses even with images in the wild. 
	We also observe that a carefully designed information source for the discriminator is essential to boost the performance. 
	Thus, we design a geometric descriptor, which computes the pairwise relative locations and distances between body joints, as a new information source for the discriminator. 
	The efficacy of our adversarial learning framework with the new geometric descriptor has been demonstrated through extensive experiments on widely used public benchmarks. 
	Our approach significantly improves the performance compared with previous state-of-the-art approaches. 
\end{abstract}

\section{Introduction}

\begin{figure}[t]
	\begin{center}
		\includegraphics[width=1\linewidth]{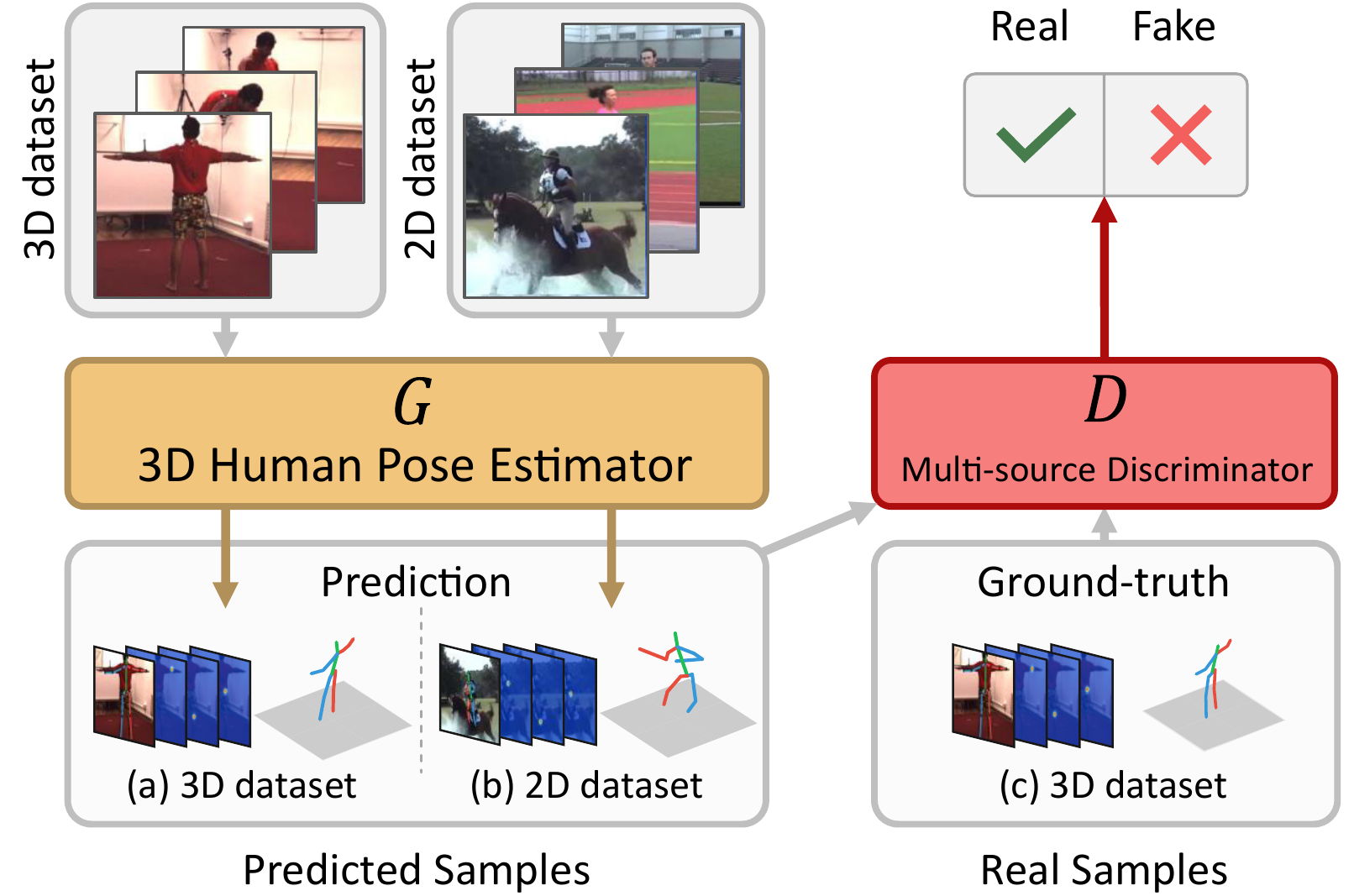}
	\end{center}
	\vspace{-0.5em}
	\caption{Given a monocular image and its predicted 3D pose, the human can easily tell whether the prediction is anthropometrically plausible or not  (as shown in b) based on the perception of image-pose correspondence and the possible human poses constrained by articulation. 
	We simulate this human perception by proposing an adversarial learning framework, where the discriminator is learned to distinguish ground-truth poses (c) from the predicted poses generated by the pose estimator (a, b), which in turn is enforced to generate plausible poses even on unannotated in-the-wild data. 
	}
	\label{fig:intro}
	\vspace{-1em}
\end{figure}

Human pose estimation is a fundamental yet challenging problem in computer vision. 
The goal is to estimate 2D or 3D locations of body parts given an image or a video, which provides informative knowledge for tasks such as action recognition, robotics vision, human-computer interaction, and autonomous driving. 
Significant advances have been achieved in 2D human pose estimation recently because of the powerful Deep Convolutional Neural Networks (DCNNs) and the availability of large-scale in-the-wild human pose datasets with manual annotations. 

However, advances in 3D human pose estimation remain limited. 
The reason is mainly from the difficulty to obtain ground-truth 3D body joint locations in the unconstrained environment. 
Existing datasets such as Human3.6M~\cite{ionescu2014human36m} are collected in the constrained lab environment using mocap systems, hence the variations in background, viewpoint, and lighting are very limited. 
Although DCNNs fit well on these datasets, when being applied on in-the-wild images, where only 2D ground-truth annotations are available (\eg, the MPII human pose dataset~\cite{andriluka20142d}), they may have difficulty in terms of generalization ability due to the large \textit{domain shift}~\cite{tzeng2017adversarial} between the constrained lab environment images and unconstrained in-the-wild images, as shown in Figure~\ref{fig:intro}.


On the other hand, given a monocular in-the-wild image and its corresponding predicted 3D pose, it is relatively easy for the human to tell if this estimation is correct or not, as demonstrated in Figure~\ref{fig:intro}(b). 
Human makes such decisions mainly based on the human perception of image-pose correspondence and possible human poses constrained by articulation. 
This human perception can be simulated by a discriminator, which is a neural network that discriminates ground-truth poses from estimations.


Based on the above observation, we propose an adversarial learning paradigm to distill the 3D human pose structures learned from the fully annotated constrained 3D pose dataset to in-the-wild images without 3D pose annotations. 
Specifically, we adopt an state-of-the-art 3D pose estimator~\cite{zhou2017towards} as a conditional generator for generating pose estimations conditioned on input images. 
The discriminator aims at distinguishing ground-truth 3D poses from predicted ones. Through adversarial learning, the generator learns to predict 3D poses that is difficult for the discriminator to distinguish from the ground-truth poses. 
Since the predicted poses can be also generated from in-the-wild data, the generator must predict indistinguishable poses on both domains to minimize the training error. 
It provides a way to train the generator, \ie, the 3D pose estimator, with in-the-wild data in a weakly supervised manner, and leads a better generalization ability. 

To facilitate the adversarial learning, a multi-source discriminator is designed to take the two key factors into consideration: 1) the description on image-pose correspondence, and 2) the human body articulation constraint. 
One indispensable information source is the original images. 
It provides rich visual information for pose-image correspondence.  
Another information source of the discriminator is the relative offsets and distances between pairs of body parts, which is motivated by traditional approaches based on pictorial structures~\cite{fischler1973representation,yang2011articulated,chen2014articulated,pishchulin2013poselet}. 
This information source provides the discriminator with rich domain prior knowledge, which helps the generator to generalize well. 

Our approach improves the state-of-the-art both qualitatively and quantitatively. 
The main contributions are summarized as follows.
\begin{myitemize}
	\item[$\bullet$] We propose an adversarial learning framework to distill the 3D human pose structures from constrained images to unconstrained domains, 
    where the ground-truth annotations are not available. 
	Our approach allows the pose estimator to generalize well on another domain in a weakly supervised manner instead of hard-coded rules. 
	
	\item[$\bullet$] We design a novel multi-source discriminator, which uses visual information as well as relative offsets and distances as the domain prior knowledge, to enhance the generalization ability of the 3D pose estimator.
\end{myitemize}
\section{Related Work} 

\subsection{2D Human Pose Estimation} 
Conventional methods usually solved 2D human poses estimation by tree-structured models, \eg, pictorial structures~\cite{pishchulin2013poselet} and mixtures of body parts~\cite{yang2011articulated,chen2014articulated}. 
These models consist of two terms: a unary term to detect the body joints, and a pairwise term to model the pairwise relationships between two body joints. 
In~\cite{yang2011articulated,chen2014articulated}, a pairwise term was designed as the relative locations and distances between pairs of body joints. 
The symmetry of appearance between limbs was modeled in~\cite{ren2005recovering,tian2010fast}. 
Ferrari \etal~\cite{ferrari20092d} designed repulsive edges between opposite-sided arms to tackle the double counting problem. 
Inspired by aforementioned works, we also use the relative locations and distances between pairs of body joints. 
But they are used as the geometric descriptor in the adversarial learning paradigm for learning better 3D pose estimation features. 
The geometric descriptor greatly reduces the difficulty for the discriminator in learning domain prior knowledge such as relative limbs length and symmetry between limbs.

Recently, impressive advances have been achieved by DCNNs~\cite{toshev2014deeppose,wei2016convolutional,newell2016stacked,chu2016structured,cao2016realtime,yang2016end,chu2017multi,yang2017learning,zhao2018pose}. 
Instead of directly regressing coordinates~\cite{toshev2014deeppose}, recent state-of-the-art methods used \textit{heatmaps}, which are generated by a 2D Gaussian centered on the body joint locations, as the target of regression. 
Our approach uses the state-of-the-art stacked hourglass~\cite{newell2016stacked} as our backbone architecture.

\subsection{3D Human Pose Estimation} 
Significant progress has been achieved for 3D human pose estimation from monocular images due to the availability of large-scale dataset~\cite{bogo2016keep} and the powerful DCNNs. 
These methods can be roughly grouped into two categories.  

One-stage approaches directly learn the 3D poses from monocular images. 
The pioneer work~\cite{li20143d} proposed a multi-task framework that jointly trains pose regression and body part detectors. 
To model high-dimensional joint dependencies, Tekin \etal~\cite{tekin2016structured} further adopted an auto-encoder at the end of the network. 
Instead of directly regressing the coordinates of the joints, Pavlakos\etal~\cite{pavlakos2016coarse} proposed a voxel representation for each joint as the regression target, and designed a coarse-to-fine learning strategy. 
These methods heavily depend on fully annotated datasets, and cannot benefit from large-scale 2D pose datasets.

Two-stage approaches first estimate 2D poses and then lift 2D poses to 3D poses~\cite{zhou2016sparseness,chen20163d,bogo2016keep,wu2016single,moreno20163d,tome2017lifting,martinez2017simple,zhou2017towards,nie2017monocular}. 
These approaches usually generalize better on images in the wild, since the first stage can benefit from the state-of-the-art 2D pose estimators, which can be trained on images in the wild. 
The second stage usually regresses the 3D locations from the 2D predictions. 
For example, Martinez~\etal~\cite{martinez2017simple} proposed a simple fully connected residual networks to directly regression 3D coordinates from 2D coordinates. 
Moreno-Noguer~\cite{moreno20163d} learned a pairwise distance matrix, which is invariant to image rotation, translation, and reflections, from 2D to 3D space. 

To predict 3D poses for images in the wild, a geometric loss was proposed in~\cite{zhou2017towards} to allow weakly supervised learning of the depth regression module. \cite{mehta2017monocular} adopted transfer
learning to generalize to in-the-wild scenes. 
\cite{mehta2017vnect} built a real-time 3D pose estimation solution with kinematic skeleton fitting. 
Our framework can use existing 3D pose estimation approaches as the baseline and is complementary to previous works by introducing an adversarial learning framework, in which the predicted 3D poses from in-the-wild images are used for learning better 3D pose estimator.

\subsection{Adversarial Learning Methods }

\smalltitle{Adversarial learning for discriminative tasks. } 
Adversarial learning has been proven effective not only for generative tasks~\cite{goodfellow2014generative,radford2015unsupervised,vondrick2016generating,zhu2017unpaired,WangECCV2016,denton2015deep,huang2017stacked,zhang2016stackgan,karras2017progressive,isola2017image,villegas2017learning,liang2017recurrent}, but also for discriminative tasks~\cite{wang2017fast,wei2017object,chen2017adversarial, chen2017adversarial2,Sohn17}. 
For example, Wang~\etal~\cite{wang2017fast} proposed to learn an adversarial network that generates hard examples with occlusions and deformations for object detection. 
Wei~\etal~\cite{wei2017object} designed an adversarial erasing approach for weakly semantic segmentation. 
An adversarial network was proposed in~\cite{chen2017adversarial, chen2017adversarial2} to distinguish the ground-truth poses from the fake ones for human pose estimation. 
The motivation and problems we are trying to tackle are completely different from these work. 
In~\cite{chen2017adversarial, chen2017adversarial2}, the adversarial loss is used to improve pose estimation accuracy with the same domain of the data. In our case, we are trying to use adversarial learning to distill the structures learned from the constrained data (with labels) in lab environments to the unannotated data in the wild. 
Our approach is also very different. \cite{chen2017adversarial, chen2017adversarial2} only trained the models in one single domain dataset, but ours incorporates the unannotated data into the learning process, which takes a large step in bridging the gap between the following two domains: 1) in-the-wild data without 3D ground-truth annotations and 2) constrained  data with 3D ground-truth annotations. 


\smalltitle{Adversarial learning for domain adaptation. }
Recently, adversarial methods have become an increasingly popular incarnation for domain adaptation tasks~\cite{ganin2015unsupervised,tzeng2015simultaneous,liu2016coupled,tzeng2017adversarial,hoffman2017cycada}. 
These methods use adversarial learning to distinguish source domain samples from target domain samples. 
And adversarial learning aims at obtaining features that are domain uninformative. 
Different from these methods, our discriminator aims at discriminating ground-truth 3D poses from the estimated ones, which can be generated either from the same domain as the ground-truth, or an unannotated domain (\eg, images in the wild). 

\section{Framework} ~\label{sec:framework}
As illustrated in Figure~\ref{fig:intro}, our proposed framework can be formulated as the Generative Adversarial Networks (GANs), which consist of two networks: a generator and a discriminator. 
The generator is trained to generate samples in a way that confuses the discriminator, which in turn tries to distinguish them from real samples. 
In our framework, the generator $G$ is a 3D pose estimator, which tries to predict accurate 3D poses to fool the discriminator. 
The discriminator $D$ distinguishes the ground-truth 3D poses from the predicted ones.  
Since the predicted poses can be generated from both the images captured from the lab environment (with 3D annotations) and unannotated images in the wild, the human body structures learned from 3D dataset can be adapted to in-the-wild images through adversarial learning.

During training, we first pretrain the pose estimator $G$ on 3D human pose dataset. 
Then we alternately optimize the generator $G$ and the discriminator $D$. 
For testing, we simply discard the discriminator.

\subsection{Generator: 3D Pose Estimator}  
The generator can be viewed as a two-stage pose estimator. 
We adopt the state-of-the-art architecture~\cite{zhou2017towards} as our backbone network for 3D human pose estimation.

The first stage is the 2D pose estimation module, which is the stacked hourglass network~\cite{newell2016stacked}. 
Each stack is in an encoder-decoder structure. 
It allows for repeated top-down, bottom-up inference across scales with intermediate supervision attached to each stack. 
We follow the previous practice to use $256\times 256$ as input resolution. 
The outputs are $P$ heatmaps for the 2D body joint locations, where $P$ denotes the number of body joints. 
Each heatmap has size $64 \times 64$. 

The second stage is a depth regression module, which consists of several residual modules taking the 2D body joint heatmaps and intermediate image features generated from the first stage as input. 
The output is a $P\times 1$ vector denoting the estimated depth for each body joint. 

A geometric loss is proposed in~\cite{zhou2017towards} to allow weakly supervised learning of the depth regression module on images in the wild. 
We discard the geometric loss for a more concise analysis of the proposed adversarial learning, although our method is complementary to theirs. 

\subsection{Discriminator}

\begin{figure}[t]
	\begin{center}
		\includegraphics[width=0.9\linewidth]{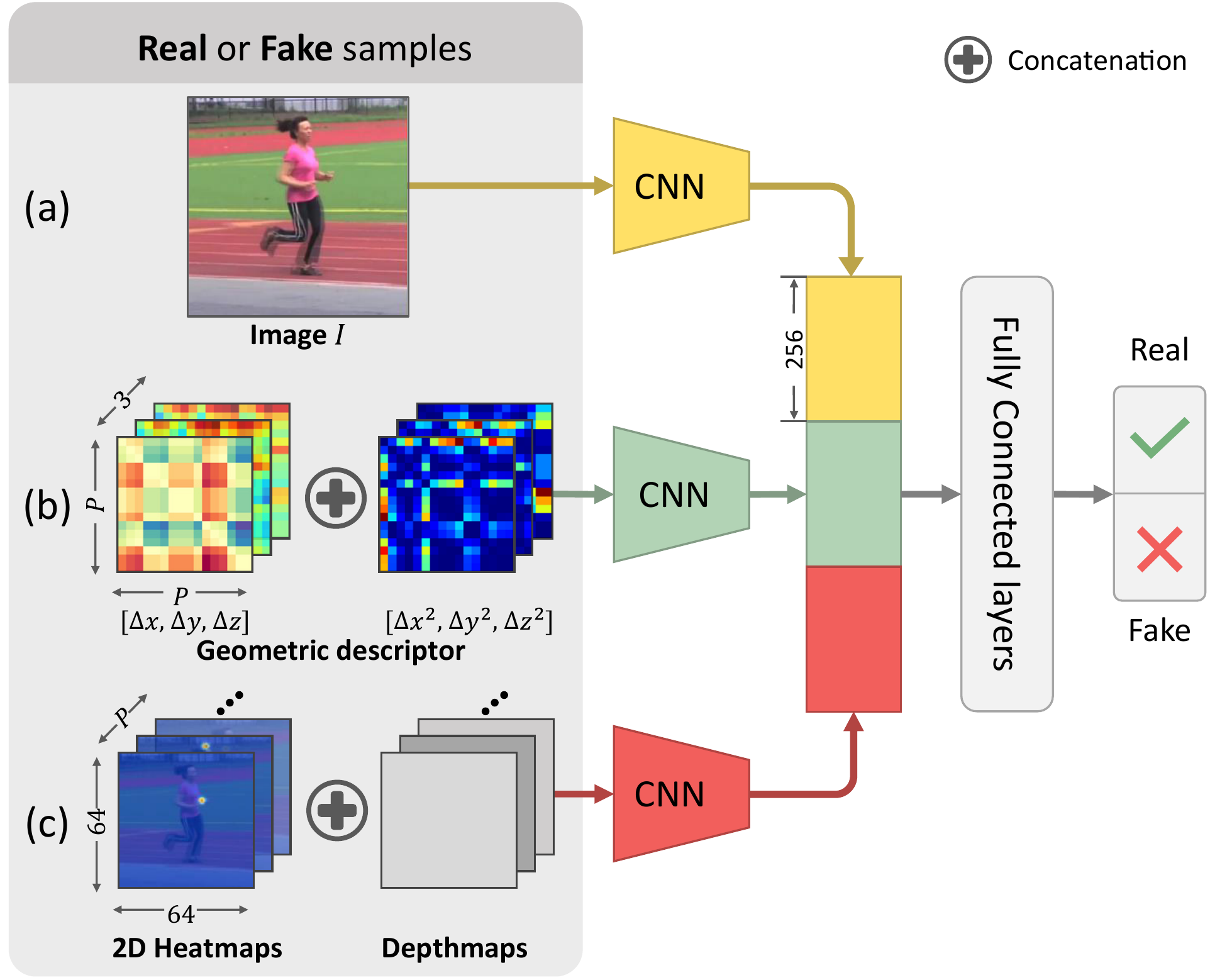}
	\end{center}
	\vspace{-1em}
	\caption{The multi-source architecture. It contains three information sources, image, geometric descriptor, as well as the heatmaps and depth maps. The three information sources are separately embedded and then concatenated for deciding if the input is the ground-truth pose or the estimated pose.  }
	\vspace{-1em}
	\label{fig:pose-d}
\end{figure}

%

The predicted poses by the generator $G$ from both the 3D pose dataset and the in-the-wild images are treated as ``fake" examples for training the discriminator $D$. 

At the adversarial learning stage, the pose estimator (generator $G$) is learned so that the ground-truth 3D poses and the predicted ones are indistinguishable for the discriminator $D$. 
Therefore, this adversarial learning enforces the predictions from in-the-wild images to have similar distributions with the ground-truth 3D poses. 
Although unannotated in-the-wild images are difficult to be directly used for training the pose estimator $G$, their corresponding 3D poses predictions can be utilized as ``fake" examples for learning better discriminator, which in turn is helpful for learning a better pose estimator (generator).

Discriminator decides whether the estimated 3D poses are similar to ground-truth or not. 
The quality of discriminator influences the pose estimator. 
Therefore, we design a multi-source network architecture and a geometric descriptor for the discriminator. 

\vspace{-1em}
\subsubsection{Multi-Source Architecture} 


In the discriminator, there are three information sources: 1) the original image, 2) the pairwise relative locations and distances, and 3) the heatmaps of 2D locations and the depths of body joints.
The information sources take two key factors into consideration: 1) the description  on image-pose  correspondence; and 2) the human body articulation constraints.

To model image-pose correspondence, we treat the original image as the first information source, which provides rich visual and contextual information to reduce ambiguities, as shown in Figure~\ref{fig:pose-d}(a).

To learn the body articulation constraints, we design a geometric descriptor as the second information source (Figure~\ref{fig:pose-d}(b)), which is motivated by traditional approaches based on pictorial structures. 
It explicitly encodes the pairwise relative locations and distances between body parts, and reduces the complexity to learn domain prior knowledge, \eg, relative limbs length, limits of joint angles,  and symmetry of body parts. 
Details are given in Section \ref{sec:geo-feats}. 

Additionally, we also investigate using heatmaps as another information source, which is effective for 2D adversarial pose estimation~\cite{chen2017adversarial}. 
It can be considered as a representation of raw body joint locations, from which the network could extract rich and complex geometric relationships within the human body structure. 
Originally, heatmaps are generated by a 2D Gaussian centered on the body part locations. 
In order to incorporate the depth information into this representation, we created $P$ depth maps, which have the same resolution as the 2D heatmaps for body joints. 
Each map is a matrix denoting the depth of a body joint at the corresponding location.   
The heatmaps and depth maps are further concatenated as the third information source, as shown in Figure~\ref{fig:pose-d} (c). 



%


\subsubsection{Geometric Descriptor}\label{sec:geo-feats} 
Our design of the geometric descriptor is motivated by the quadratic deformation constraints widely used in pictorial structures~\cite{yang2011articulated,pishchulin2013poselet,chen2014articulated} for 2D human pose estimation. 
It encodes the spatial relationships, limbs length and symmetry of body parts.  
By extending it from 2D to 3D space, we define the 3D geometric descriptor $d(\cdot, \cdot)$ between pairs of body joints as a 6D vector 
\begin{eqnarray}
d(\bz_i, \bz_j) = [\Delta x, \Delta y, \Delta z, \Delta x^2, \Delta y^2, \Delta z^2]^T,
\label{eq:geo_des}
\end{eqnarray}
where $\bz_i=(x_i, y_i, z_i)$ and $ \bz_j=(x_j, y_j, z_j)$ denote the 3D coordinates of the body joint $i$ and $j$. 
$\Delta x = x_i - x_j, \Delta y = y_i - y_j $ and $\Delta z = z_i - z_j$ are the relative locations of joint $i$ with respect to joint $j$. 
$ \Delta x^2=(x_i - x_j)^2, \Delta y^2=(y_i - y_j)^2$ and $ \Delta z^2=(z_i - z_j)^2$ are distances between $i$ and $j$. 

We compute the $6$D geometric descriptor  in Eq. (\ref{eq:geo_des}) for each pair of body joint, which results in a $6\times P \times P$ matrix  for $P$ body joints. 

\section{Learning}
GANs are usually trained from scratch by optimizing the generator and the discriminator alternately~\cite{goodfellow2014generative,radford2015unsupervised}. 
For our task, however, we observe that the training will converge faster and get better performance with a pretrained generator (\ie, the 3D pose estimator). 

We first briefly introduce the notation. 
Let $ \I = \{ (I_n, \bz_n)\}_{n=1}^N $ denote the datasets, 
where $N$ denote the sample indexes. 
Specifically,  $N = \{N_{2D}, N_{3D}\}$, where $N_{2D}$ and $ N_{3D}$ are sample indexes for the 2D and 3D pose datasets. 
Each sample $ (I, \bz) $  consists of a monocular image $I$ and the ground-truth body joint locations $\bz$, 
where $\bz = \{(x^j, y^j)\}_{j=1}^P$ for 2D pose dataset, and $\bz = \{(x^j, y^j, z^j)\}_{j=1}^P$ for 3D pose dataset. 
Here $P$ denote the number of body joints. 

\subsection{Pretraining of the Generator} 

We first pretrain the 3D pose estimator (\ie the generator), which consists of the 2D pose estimation module and the depth regression module. 
We follow the standard pipeline~\cite{tompson2015efficient,wei2016convolutional,cao2016realtime,newell2016stacked} and formulate the 2D pose estimation as the heatmap regression problem. 
The ground-truth heatmap $\bS^j$ for body joint $j$ is generated from a Gaussian centered at $(x^j, y^j)$ with variance $\mathbf{\Sigma}$, which is set as an identity matrix empirically. 
Denote the predicted 2D heatmaps and depth as $\hat{\bS}^j$ and $\hat{z}^j$ respectively. 
The overall loss for training pose estimator is defined as the squared error
{\small
\begin{eqnarray}
	\mathcal{L}_{pose} = \sum_{j=1}^{P} \left( \sum_{n\in N}  \overbrace{ \| \bS_n^j -  \hat{\bS}_n^j\|^2_2}^\text{heatmap regression} 
	+ \sum_{n\in N_{3D}}   \overbrace{ \| z_n^j -  \hat{z}_n^j\|^2_2}^\text{depth regression} \right).
	 \label{eq:loss-pose} 
\end{eqnarray}
}

As in previous works~\cite{martinez2017simple,zhou2017towards}, we adopt a pretrained stacked hourglass networks~\cite{newell2016stacked} as the 2D pose estimation module. 
Then the 2D pose module and the depth regression module are jointly fine-tuned with the loss in Eq.(\ref{eq:loss-pose}).

\subsection{Adversarial Learning} 

\begin{figure}[t]
	\begin{center}
		\includegraphics[width=1\linewidth]{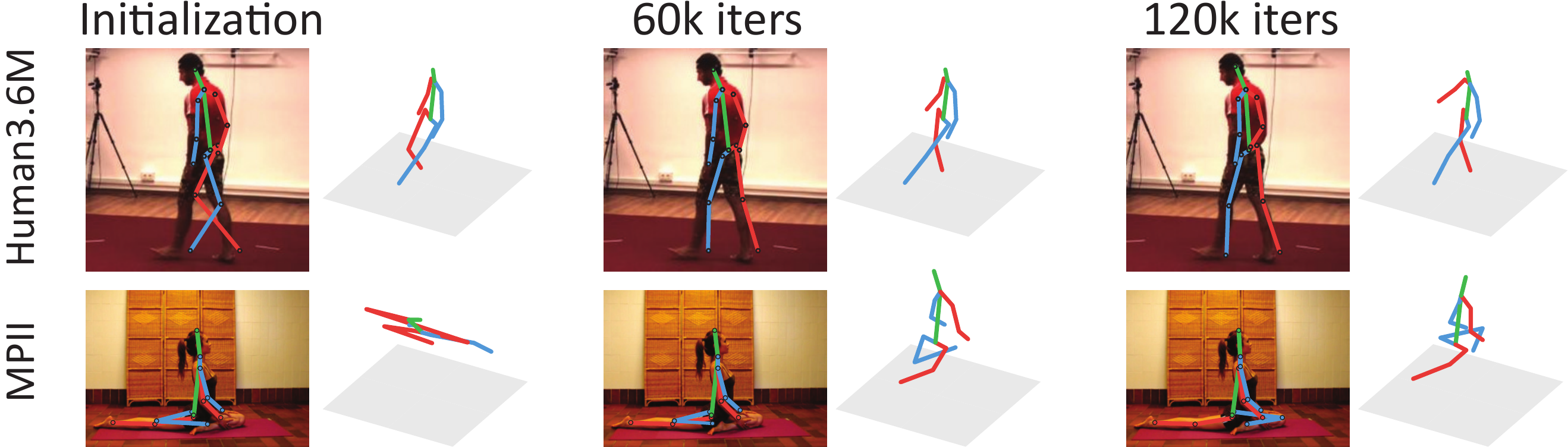}
	\end{center}
	\caption{The predicted 3D poses become more accurate along with the adversarial learning process.}
	\label{fig:pose3d-adversary}
\end{figure}

After pretraining the 3D pose estimator $G$, we alternately optimize $G$ and $D$. 
The loss for training discriminator $D$ is, 
{\small
\begin{eqnarray}
	\mathcal{L}_D  =  \sum_{n\in N_{3D}} \mathcal{L}_{cls}\left(D(I_n, E(\bS_n, \bz_n)), 1\right) \nonumber \\
	+ \sum_{n\in N} \mathcal{L}_{cls}(D\left(I_n, E(G(I_n))), 0\right), 
	\label{eq:loss-d}
\end{eqnarray}
}
\!\!\!\!where $E(\bS_n, \bz_n)$ encodes the heatmaps, depth maps and the geometric descriptor as described in Section~\ref{sec:framework}. 
$D(I_n, E(\bS_n, \bz_n))\in [0, 1]$ represents the classification score of the discriminator given input image $I_n$ and the encoded information  $E(\bS_n, \bz_n)$. 
$G(I_n)$ is a 3D pose estimator which predicts heatmaps $\hat{\bS}_n^j$ and depth values $\hat{z}_n^j$ given an input image $I_n$. 
$L_{cls}$ is the binary entropy loss defined as $\mathcal{L}_{cls}(\hat{y}, y) = - (y\log(\hat{y}) + (1-y)\log(1-\hat{y}))$. 
Within each minibatch, half of samples are ``real" from the 3D pose dataset, and the rest $(I_n, E(G(I_n)))$ are generated by
$G$ given an image $I_n$ from 3D or 2D pose dataset. 
Intuitively, $L_D$ is optimized to enforce the network $D$ to classify the ground-truth poses as label 1 and the predictions as 0. 

On the contrary, the generator $G$ tries to generalize anthropometrically plausible poses conditioned on an image to fool $D$ via minimizing the following classification loss,
\begin{eqnarray}
\mathcal{L}_G  = \sum_{n\in N} \mathcal{L}_{cls}(D\left(I_n, E(G(I_n))), 1\right).
\label{eq:loss-g1}
\end{eqnarray}
We observe that directly train $G$ and $D$ with the loss proposed in Eq.(\ref{eq:loss-d}) and Eq.(\ref{eq:loss-g1}) reduces the accuracies of the predicted poses. 
To regularize the training process, we incorporate the regression loss $L_{pose}$ in Eq.(\ref{eq:loss-pose}) into Eq.(\ref{eq:loss-g1}), which results in the following loss function,
\begin{eqnarray}
\mathcal{L}_G  = \lambda\sum_{n\in N} \mathcal{L}_{cls}(D\left(I_n, E(G(I_n))), 1\right) + \mathcal{L}_{pose},
\label{eq:loss-g}
\end{eqnarray}
where $\lambda$ is a hyperparameter to adjust the trade-off between the classification loss and the regression loss. 
$\lambda$ is set as $1e-4$ in the experiments.

Figure~\ref{fig:pose3d-adversary} demonstrates the improvements of predicted 3D poses with the adversarial learning process. 
The initial predictions are anthropometrically invalid, and are easily distinguishable by $D$ from the ground-truth poses. 
A relatively large error $\mathcal{L}_G$ is thus generated, and $G$ is updated accordingly to fool $D$ better and produce improved results.


\begin{table*}
	\centering
	\setlength{\tabcolsep}{4pt}
	\resizebox{\textwidth}{!}{
		\begin{tabular}{@{}lccccccccccccccc|c@{}}
			\toprule
			\textbf{Protocol \#1} & Direct. & Discuss & Eating & Greet & Phone & Photo & Pose & Purch. & Sitting & SittingD. & Smoke & Wait & WalkD. & Walk & WalkT. & Avg.\\
			\midrule
			LinKDE PAMI'16~\cite{ionescu2014human36m} & 132.7 & 183.6 & 132.3 & 164.4 & 162.1 & 205.9 & 150.6 & 171.3 & 151.6 & 243.0 & 162.1 & 170.7 & 177.1 & 96.6 & 127.9 & 162.1\\
			Tekin~\etal, ICCV'16~\cite{tekin2016direct} & 102.4 & 147.2 & 88.8 & 125.3 & 118.0 & 182.7 & 112.4 & 129.2 & 138.9 & 224.9 & 118.4 & 138.8 & 126.3 & 55.1 & 65.8 & 125.0\\
			Du~\etal ECCV'16~\cite{du2016marker} & 85.1 & 112.7 & 104.9 & 122.1 & 139.1 & 135.9 & 105.9 & 166.2 & 117.5 & 226.9 & 120.0 & 117.7 & 137.4 & 99.3 & 106.5 & 126.5\\
			Chen \& Ramanan CVPR'17~\cite{chen20163d} & 89.9 & 97.6 & 89.9 &  107.9 &  107.3 & 139.2 &  93.6 &  136.0 & 133.1 & 240.1 & 106.6 &  106.2 &  87.0 & 114.0 & 90.5 & 114.1 \\
			Pavlakos~\etal CVPR'17~\cite{pavlakos2016coarse} & 67.4 & 71.9 & 66.7 & 69.1 & 72.0 & 77.0 & 65.0 & 68.3 & 83.7 & 96.5 & 71.7 & 65.8 & 74.9 & 59.1 & 63.2 & 71.9\\
			Mehta~\etal 3DV'17~\cite{mehta2017monocular} & 52.6 & 64.1 & 55.2 & 62.2 &  71.6 & 79.5 &  52.8 &  68.6 & 91.8 & 118.4 & 65.7 &  63.5 &  49.4 & 76.4 & 53.5 & 68.6 \\
			Zhou~\etal. ICCV'17~\cite{zhou2017towards} & 54.8 & 60.7 & 58.2 & 71.4 & 62.0 & 65.5 & 53.8 & 55.6 & 75.2 & 111.6 & 64.1 & 66.0 & 51.4 & 63.2 & 55.3 & 64.9\\
			Martinez~\etal ICCV'17~\cite{martinez2017simple} & 51.8& 56.2& 58.1&	59.0& 69.5&	78.4& 55.2&	58.1&74.0&94.6&	62.3&59.1&65.1&49.5&52.4&62.9\\
			Fang~\etal AAAI'18~\cite{fang2017learning} & 50.1&  54.3&	57.0&	57.1&	66.6& 73.3&	53.4&	55.7&	72.8&	88.6&	60.3&	57.7&	62.7&	47.5&	50.6&	{60.4}\\
			\midrule
			Ours (Full-2s) & 53.0 & 60.8 & 47.9 & 57.1 & 61.5 & 65.5 & 50.8 & 49.9 & 73.3 & 98.6 & 58.8 & 58.1 & 42.0 & 62.3 & 43.6  & 59.7 \\
			Ours (Full-4s) & 51.5 & 58.9 & 50.4 & 57.0 & 62.1 & 65.4 & 49.8 & 52.7 & 69.2 & 85.2 & 57.4 & 58.4 & 43.6 & 60.1 & 47.7 & \bf{58.6} \\
			\toprule
			\textbf{Protocol \#2} & Direct. & Discuss & Eating & Greet & Phone & Photo & Pose & Purch. & Sitting & SittingD. & Smoke & Wait & WalkD. & Walk & WalkT. & Avg.\\
			\midrule
			Ramakrishna~\etal ECCV'12~\cite{ramakrishna2012reconstructing} & 137.4 & 149.3 & 141.6 & 154.3 & 157.7 & 158.9 & 141.8 & 158.1 & 168.6 & 175.6 & 160.4 & 161.7 & 150.0 & 174.8 & 150.2 & 157.3\\
			Bogo~\etal ECCV'16~\cite{bogo2016keep} & 62.0 & 60.2 & 67.8 & 76.5 & 92.1 & 77.0 & 73.0 & 75.3 & 100.3 & 137.3 & 83.4 & 77.3 & 86.8 & 79.7 & 87.7 & 82.3\\
			Moreno-Noguer CVPR'17~\cite{moreno20163d} & 66.1 & 61.7 & 84.5 & 73.7 & 65.2 & 67.2 & 60.9 & 67.3 & 103.5 & 74.6 & 92.6 & 69.6 & 71.5 & 78.0 & 73.2 & 74.0\\
			Pavlakos~\etal CVPR'17~\cite{pavlakos2016coarse}  & -- & -- & -- & -- & -- & -- & -- & -- & -- & -- & -- & -- & -- & -- & -- & 51.9\\
			Martinez~\etal ICCV'17~\cite{martinez2017simple} & 39.5 & 43.2&46.4&	47.0&	51.0&	56.0&	41.4&	40.6&	56.5&	69.4&	49.2&	45.0&	49.5&	38.0&	43.1&	47.7\\
			Fang~\etal AAAI'18~\cite{fang2017learning} & 38.2 & 41.7&	43.7&	44.9&	48.5&	55.3&	40.2&	38.2&	54.5&	64.4&	47.2&	44.3&	47.3&	36.7&	41.7&	45.7\\
			\midrule
			Ours (Full-4s) & 26.9 & 30.9 & 36.3 & 39.9 & 43.9 & 47.4 & 28.8 & 29.4 & 36.9 & 58.4 & 41.5 & 30.5 & 29.5 & 42.5 & 32.2 & \bf{37.7}
			\\
			\bottomrule
		\end{tabular}
	}
	\caption{Quantitative comparisons of Mean Per Joint Position Error (MPJPE) in millimetre between the estimated pose and the ground-truth on \textit{Human3.6M} under \textit{Protocol \#1} and \textit{Protocol \#2}. Some results are borrowed from~\cite{fang2017learning}. }
	\vspace{-1em}
	\label{tab:h36m}
\end{table*}

\section{Experiments}
\smalltitle{Datasets.} 
We conduct experiments on three popular human pose estimation benchmarks: Human3.6M~\cite{ionescu2014human36m}, MPI-INF-3DHP~\cite{mehta2017monocular} and MPII Human Pose~\cite{andriluka20142d}. 

Human3.6M~\cite{ionescu2014human36m} dataset is one of the largest datasets for 3D human pose estimation. 
It consists of 3.6 million images featuring 11 actors performing 15 daily activities, such as eating, sitting, walking and taking a photo, from 4 camera views.
The ground-truth 3D poses are captured by the Mocap system, while the 2D poses can be obtained by projection with the known intrinsic and extrinsic camera parameters. 
We use this dataset for quantitative evaluation. 

MPI-INF-3DHP~\cite{mehta2017monocular} is a recently proposed 3D dataset constructed by the Mocap system with both constrained indoor scenes and complex outdoor scenes. We only use the test split of this dataset, which contains 2929 frames from six subjects performing seven actions, to evaluate the generalization ability quantitatively. 

The MPII Human Pose~\cite{andriluka20142d} is the standard benchmark for 2D human pose estimation. 
It contains 25K unconstrained images collected from YouTube videos covering a wide range of activities. 
We adopt this dataset for the 2D pose estimation evaluation and the qualitative evaluation.

\smalltitle{Evaluation protocols.}  
We follow the standard protocol on Human3.6M to use the subjects 1, 5, 6, 7, 8 for training and the subjects 9 and 11 for evaluation. 
The evaluation metric is the Mean Per Joint Position Error (MPJPE) in millimeter between the ground-truth and the prediction across all cameras and joints after aligning the depth of the root joints.
We refer to this as \textit{Protocol \#1}.
In some works, the predictions are further aligned with the ground-truth via a rigid transform~\cite{bogo2016keep, moreno20163d, martinez2017simple}, which is referred as \textit{Protocol \#2}.

\smalltitle{Implementation details.}  
We adopt the network architecture proposed in~\cite{zhou2017towards} as the backbone of our pose estimator. 
Specifically, for 2D pose module, we adopt a shallower version of stacked hourglass~\cite{newell2016stacked}, \ie 2 stacks with 1 residual module at each resolution, for fast training in ablation studies (Table~\ref{tab:ablation-h36m}). 
The final results in Table~\ref{tab:h36m} are generated with 4 stacks of hourglass with 1 residual module at each resolution (\ie Ours (Full-4s)), which has approximately the same number of parameters but better performance compared with the structure (2 stacks with 2 residual module at each resolution) used in~\cite{zhou2017towards}. 
The depth regression module consists of three sequential residual and downsampling modules, a global average pooling, and a fully connected layer for regressing the depth. 
The discriminator consists of three fully connected layers after concatenating the three (or two) branches of features embedded from three information sources, \ie the image, the heatmaps and depth maps, and the pairwise geometric descriptors.  

Following the standard training procedure as in~\cite{zhou2017towards, martinez2017simple}, we first pretrain the 2D pose estimator on the MPII dataset to match the performance reported in~\cite{newell2016stacked}. Then we train the full pose estimator with the pretrained 2D module on Human3.6M for 200K iterations. 
To distill the learned 3D poses to the unconstrained dataset, we then alternately train the discriminator and pose estimator for 120k iterations. The batch size is 12 for all the steps. 
All the experiments were conducted on a single Titan X GPU. 
The forward time during testing is about $1.1$ second for a batch of 24 images.  

%

\begin{figure*}
	\begin{center}
		\includegraphics[width=0.9\linewidth]{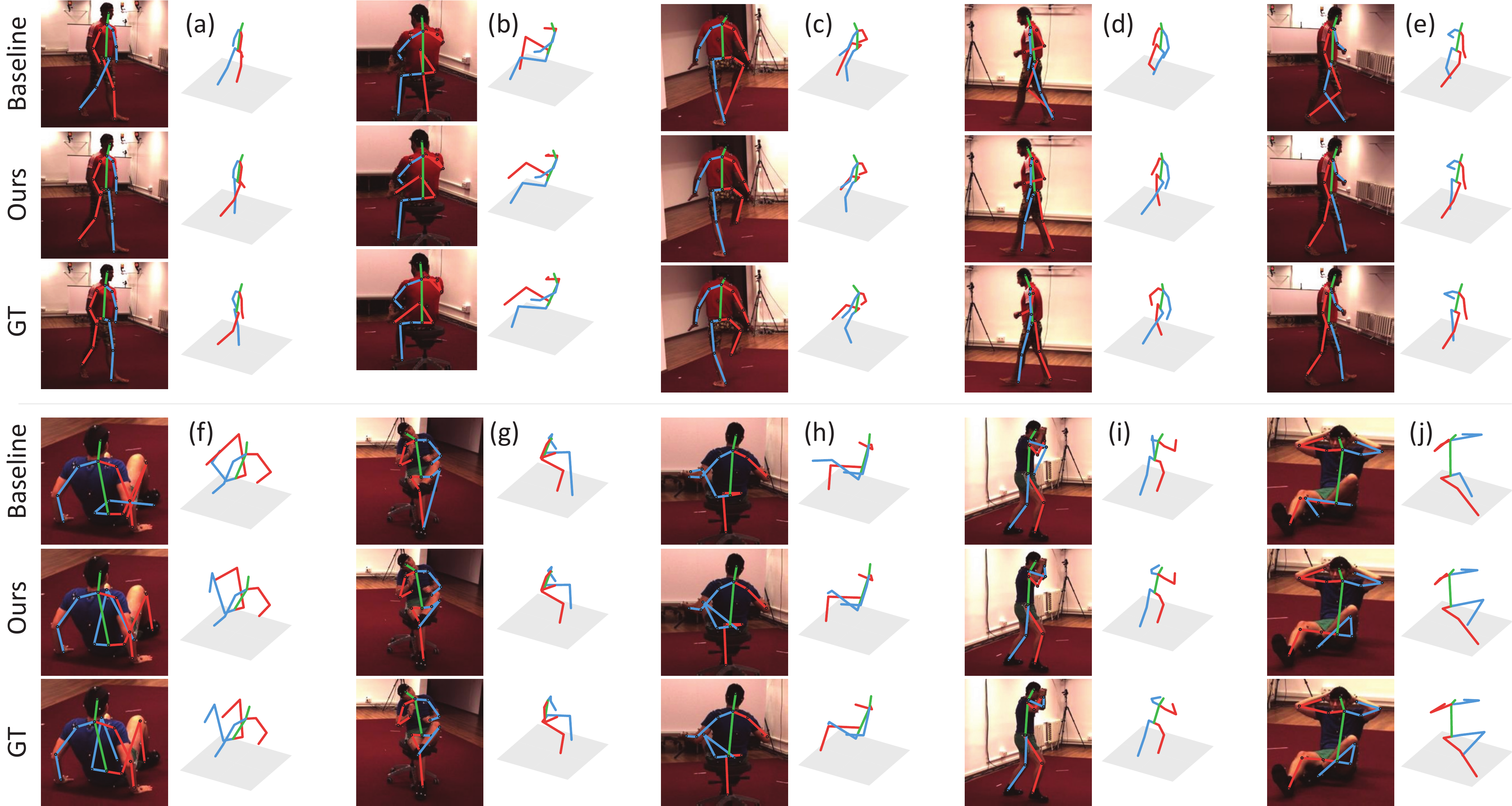}
	\end{center}
	\vspace{-1em}
	\caption{Predicted 3D poses on the Human3.6M validation set. 
		Compared with the baseline pose estimator, the proposed adversarial learning framework (\textit{Ours}) is able to refine the anatomically implausible poses, which is more similar to the ground-truth poses (\textit{GT}). }
	\vspace{-1em}
	\label{fig:h36m}
\end{figure*}

\subsection{Results on Human3.6M}
Table~\ref{tab:h36m} reports the comparison with previous methods on Human3.6M. 
Our method  (\ie Ours (Full-4s)) achieves the state-of-the-art results. 
For \textit{Protocol \#1}, our method obtains $58.6$ of mm of error, which has $9.7\%$ improvements compared to our backbone architecture~\cite{zhou2017towards}, although the geometric loss used in~\cite{zhou2017towards} is not used in our model for clearer analysis. 
Comparing to the recent best result~\cite{fang2017learning}, our method still has $3.0\%$ improvement.

Under \textit{Protocol \#2} (predictions are aligned with the ground-truth via a rigid transform), our method obtains $37.7$mm error, which improves the previous best result~\cite{fang2017learning}, $45.7$mm, on a large margin ($17.5\%$ improvement).  

\subsubsection{Ablation Study} 
To investigate the efficacy of each component, we conduct ablation analysis on Human3.6M under \textit{Protocol \#1}. 
For fast training, we adopt a shallower version of the stacked hourglass, \ie 2 stacks with 1 residual module at each resolution  (Ours (Full-2s) in Table~\ref{tab:h36m}), as the backbone architecture for the 2D pose module.  
Mean errors of all the joints and four limbs (\ie, upper/lower arms and upper/lower legs) are reported in Table~\ref{tab:ablation-h36m}. 
The notations are as follows:
\begin{myitemize}
	\item[$\bullet$] \textbf{Baseline} refers to the pose estimator without adversarial learning. The mean error of our baseline model is 64.8 mm, which is very close to the 64.9 mm error reported on our backbone architecture in~\cite{zhou2017towards}.  
	\item[$\bullet$] \textbf{Map} refers to the use of heatmaps and depth maps, as well as the original images for the adversarial training. 
	\item[$\bullet$] \textbf{Geo} refers to use our proposed geometric descriptors as well as the original images for the adversarial training.
	\item[$\bullet$] \textbf{Full} refers to use all the information sources, \ie, original images, heatmaps and depth maps, and geometric descriptors, for adversarial learning.  
	\item[$\bullet$] \textbf{Fix 2D} refers to training with 2D pose  module fixed. 
	\item[$\bullet$] \textbf{W/o pretrain} refers to adversarial learning without pretraining the depth regressor. 
\end{myitemize}


\begin{table}
	\begin{footnotesize}
		\centering
		\begin{tabular}{@{}p{2.1cm}p{0.8cm}p{0.8cm}p{0.8cm}p{0.8cm}|p{0.7cm}}
			\hline
			Method 					& U.Arms & L.Arms & U.Legs & L.Legs & Mean	\\
			\hline
			Baseline (fix 2D)		& 67.6 & 89.6 & 46.6 & 83.3 & 65.2 \\
			Baseline 				& 66.6 & 90.0 & 47.1 & 83.7 & 64.8 \\
			\hline
			Map					& 62.9 & 81.6 & 44.6 & 80.9 & 61.3 \\
			Geo				& \textbf{61.6} & \textbf{80.7} & 43.9 & 78.8 & 60.3 \\
			\hline
			Full (fix 2D)		 	& 63.9 & 84.4 & 45.8 & 85.1 & 63.1 \\
			Full (w/o pretrain)	 	& 65.2 & 84.2 & 46.7 & 82.5 & 63.4  \\
			Full					& 61.7 & 81.1 & \textbf{43.1} & \textbf{77.6} & \textbf{59.7} \\
			\hline 
		\end{tabular}
		\vspace{0.2em}
		\caption{Ablation studies on the Human3.6M dataset under \textit{Protocol \#1} with 2 stacks of hourglass. 
			The first two rows refer to the baseline pose estimator without adversarial learning. 
			Rest of the rows refer to variants with adversarial learning. 
			Please refer to the text for the detailed descriptor for each variant. }
		\label{tab:ablation-h36m}
		\vspace{-0.5em}
	\end{footnotesize}
\end{table}


\begin{figure}
	\begin{center}
		\includegraphics[width=0.95\linewidth]{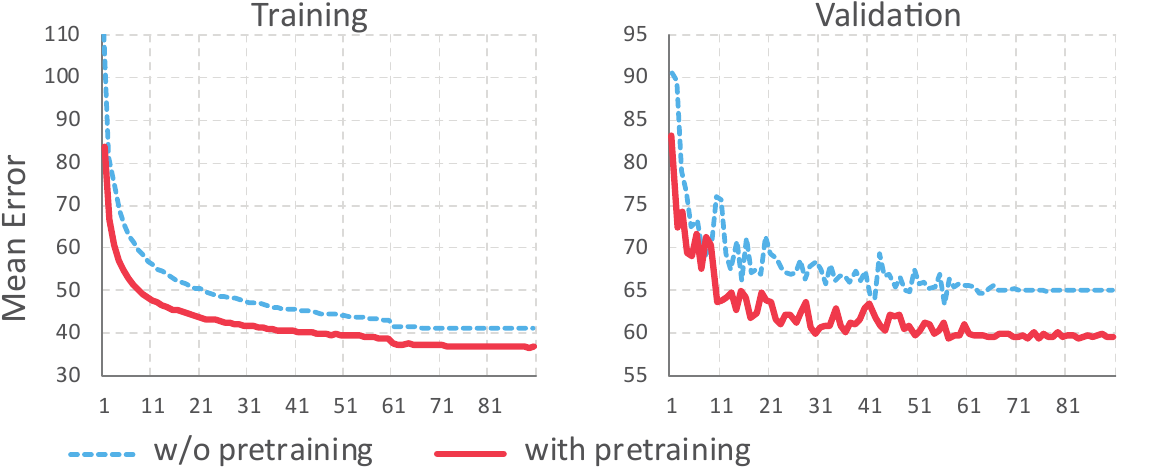}
	\end{center}
	\vspace{-1em}
	\caption{Training and validation curves of MPJPE (mm) vs. epoch
		on the Human3.6M validation set. Better convergence rate and performance have been achieved with the pretrained generator.}
	\label{fig:curve}
	\vspace{-1em}
\end{figure}

\begin{figure*}[t]
	\begin{center}
		\includegraphics[width=0.9\linewidth]{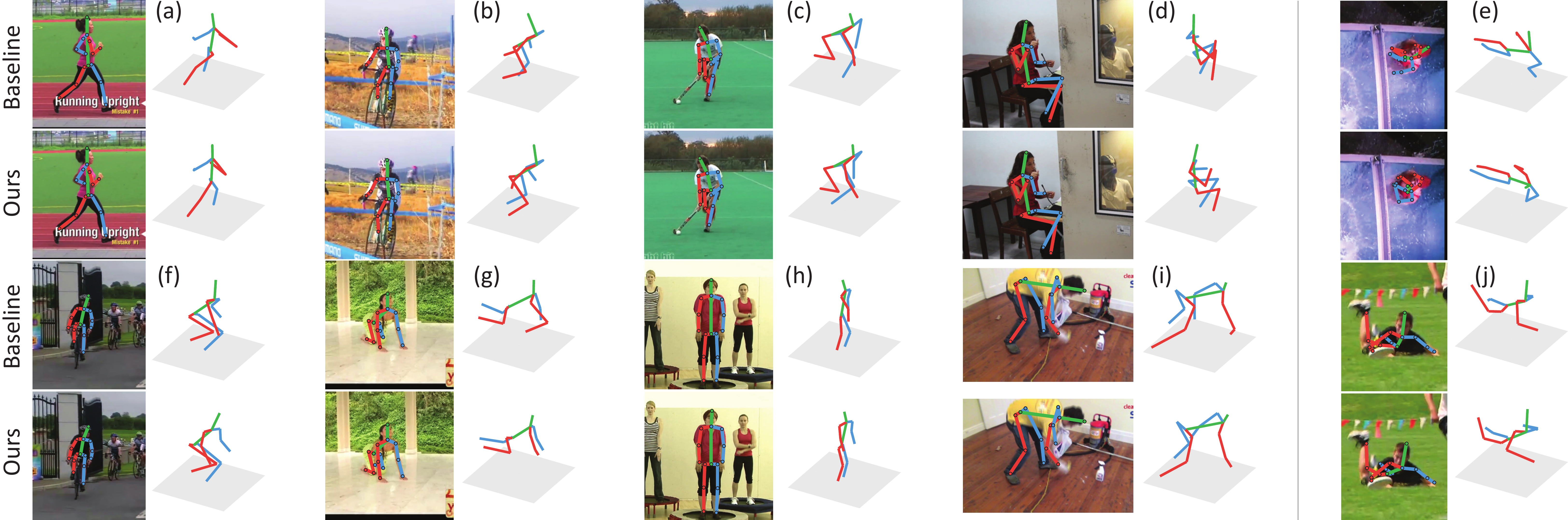}
	\end{center}
	\vspace{-1em}
	\caption{Qualitative comparison of images in the wild (\ie the MPII human pose dataset~\cite{andriluka20142d}). 
		anatomically implausible bent of limbs is corrected by the adversarial learning. 
		The last column shows typical failure cases caused by unseen camera views. 
	}
	\vspace{-1em}
	\label{fig:mpii}
\end{figure*}

\smalltitle{Geometric features: heatmaps or pairwise geometric descriptor?} 
From Table~\ref{tab:ablation-h36m}, we observe that all the variants with adversarial learning outperform the baseline model. 
If we use the image, the heatmaps and the depth maps as the information source (\textit{Map}) for the discriminator, the prediction error is reduced by 3.5 mm. 
From the baseline model, the pairwise geometric descriptor (\textit{Geo}) introduced in Section~\ref{sec:geo-feats} reduces the prediction error by 4.5 mm. The pairwise geometric descriptor provides 1 mm lower mean error compared to the heatmaps (Map).  
This validates the effectiveness of the proposed geometric features in learning complex constraints in the articulated human body. 
By combining all the three information sources together (\textit{Full}), our framework achieves the lowest error.



\smalltitle{Adversarial learning: from scratch or not? }
The standard practice to train GANs is to learn the generator and the discriminator alternately from scratch~\cite{goodfellow2014generative,radford2015unsupervised,vondrick2016generating,zhu2017unpaired}. 
The generator is usually conditioned on noise~\cite{radford2015unsupervised}, text~\cite{zhang2016stackgan} or images~\cite{zhu2017unpaired}, and lacks of ground-truth for supervised training. 
This may not be necessary for our case because our generator is actually the pose estimator and can be pretrained in a supervised manner. 
To investigate which training strategy is better, we train our full model with or without pretraining the depth regressor. 
We found that it is easier to learn when the generator is pretrained: 
It not only obtains lower prediction error (59.7 \vs 63.4 mm), but also converges much faster, as shown by the training and validation curves of mean error \vs epoch in Figure~\ref{fig:curve}. 


\begin{table}
\begin{footnotesize}
	\centering
	\begin{tabular} 
		{@{}p{1cm}|p{0.4cm}p{0.4cm}p{0.4cm}p{0.4cm}p{0.4cm}p{0.4cm}p{0.5cm}|p{0.5cm}}
		\hline
		Model & Head & Sho. & Elb. & Wri. & Hip & Knee & Ank. & Mean\\
		\hline
		Pretrain & \textbf{96.3} & 95.0   &  89.0 &  84.5 &   87.1 &  82.5 &  78.3 &   87.6  \\
		Ours & 96.1 & \textbf{95.6}  &   \textbf{89.9} & \textbf{84.6} &  \textbf{87.9} &  \textbf{84.3} & \textbf{81.2}  & \textbf{88.6} \\
		\hline
	\end{tabular}
\end{footnotesize}
\caption{PCKh@0.5 score on the MPII validation set. }
\vspace{-1em}
\label{tab:MPII}
\end{table}

\smalltitle{Shall we fix the pretrained 2D module?} 
Since the 2D pose estimator is mature enough~\cite{newell2016stacked,wei2016convolutional,cao2016realtime}. 
Is it still necessary to learn our model end-to-end to the 2D pose module with more computational and memory cost? 
We first investigate this issue with the baseline model.  
For the baseline model, the top rows of Table~\ref{tab:ablation-h36m} show that end-to-end learning (\textit{Baseline}) is similar in performance compared to the learning of depth regressor with 2D module fixed (\textit{Baseline (Fix-2D)}). 
For adversarial learning, on the other hand, the improvement from end-to-end learning is obvious, with 3.4 mm (around 5\%) error reduction when compare \textit{Full (Fix 2D)} with \textit{Full} in the table.
Therefore, end-to-end training is necessary to boost the performance in adversarial learning.

\smalltitle{Adversarial learning for 2D pose estimation. }
One may wonder the performance of 2D module after the adversarial learning. 
Therefore, we reported the PCKh@0.5 scores for 2D pose estimation on the MPII validation set in Table~\ref{tab:MPII}.  
\textit{Pretrain} refers to our baseline 2D module without adversarial training. 
\textit{Ours} refers to the the model after the adversarial learning.  
We observe that adversarial learning reduces the error rate of 2D pose estimation by $8.1\%$. 

\smalltitle{Qualitative comparison. } 
To understand how adversarial learning works, we compare the poses estimated by the baseline model to those generated with adversarial learning. 
Specifically, the high-level domain knowledge over human poses, such as symmetry (Figure~\ref{fig:h36m} (b,c,f,g,i)) and kinematics (Figure~\ref{fig:h36m} (b,c,g,f,i)), are encoded by the adversarial learning.
Hence the generator (\ie the pose estimator) is able to refine the anatomically implausible poses, which might be caused by left-right switch (Figure~\ref{fig:h36m} (a, e)), cluttered background (Figure~\ref{fig:h36m} (b)), double counting (Figure~\ref{fig:h36m} (c,d,g)) and severe occlusion (Figure~\ref{fig:h36m} (f,h,i)). 

\subsection{Cross-Domain Generalization}

\smalltitle{Quantitative results on MPI-INF-3DHP. } 
One way to show that our algorithm is learning to transfer between domains is to test our model on another unseen 3D pose estimation dataset. 
Thus, we add a cross-dataset experiment on a recently proposed 3D dataset MPI-INF-3DHP~\cite{mehta2017monocular}. 
For training, only the H36M and MPII are used, while MPI-INF-3DHP is not used. 
We follow~\cite{mehta2017monocular} to use PCK and AUC as the evaluation metrics. 
Comparisons are reported in Table~\ref{tab:mpi-inf-3dhp}. 
\textit{Baseline} and \textit{Adversarial} denote the pose estimator without or with the adversarial learning, respectively. 
We observe that the adversarial learning significantly improves the generalization ability of the pose estimator. 

\smalltitle{Qualitative results on MPII. } 
Finally, we demonstrate the generalization ability qualitatively on the validation split of the in-the-wild MPII human pose~\cite{andriluka20142d} dataset. 
Compared with the \textit{baseline} method without adversarial learning, our discriminator is able to identify the unnaturally bent limbs (Figure~\ref{fig:mpii}(a-c,g-i)) and asymmetric limbs (Figure~\ref{fig:mpii}(d)), and to refine the pose estimator through adversarial training. 

One common failure case is shown in Figure~\ref{fig:mpii}(e). 
The picture is a high-angle shot, which is not covered by the four cameras in the 3D pose dataset. 
This issue could be probably solved by involving more camera views during training. 

\begin{table}
	\centering
    \begin{footnotesize}
	\begin{tabular} {l | c  c  c}
		\hline
		 & \cite{mehta2017monocular} & Baseline & Ours \\
		\hline
		 PCK & 64.7 & 50.1 & \textbf{69.0} \\ 
		 AUC & 31.7 & 21.6 & \textbf{32.0} \\
		\hline
	\end{tabular}
	\caption{PCK and AUC on the MPI-INF-3DHP dataset. 
	}
	\vspace{-2em}
	\label{tab:mpi-inf-3dhp}
    \end{footnotesize}
\end{table}

\section{Conclusion}
This paper has proposed an adversarial learning framework to transfer the 3D human pose structures learned from the fully annotated dataset to in-the-wild images with only 2D pose annotations.  
A novel multi-source discriminator, as well as a geometric descriptor to encode the pairwise relative locations and distances between body joints, have been introduced to bridge the gap between the predicted pose from both domains and the ground-truth poses. Experimental results validate that the proposed framework improves the pose estimation accuracy on 3D human pose dataset. 
In the future work, we plan to investigate the augmentation of camera views for better generalization ability. 

\smalltitle{Acknowledgment}: 
This work is supported in part by SenseTime Group Limited, in part by the General Research Fund through the Research Grants Council of Hong Kong under Grants CUHK14213616, CUHK14206114, CUHK14205615, CUHK419412, CUHK14203015, CUHK14239816, CUHK14207814, CUHK14208417, CUHK14202217,  in part by the Hong Kong Innovation and Technology Support Programme Grant ITS/121/15FX.

{\small
\bibliographystyle{ieee}
\bibliography{egbib}

\begin{thebibliography}{10}\itemsep=-1pt

\bibitem{andriluka20142d}
M.~Andriluka, L.~Pishchulin, P.~Gehler, and B.~Schiele.
\newblock 2d human pose estimation: New benchmark and state of the art
  analysis.
\newblock In {\em CVPR}, 2014.

\bibitem{bogo2016keep}
F.~Bogo, A.~Kanazawa, C.~Lassner, P.~Gehler, J.~Romero, and M.~J. Black.
\newblock Keep it smpl: Automatic estimation of 3d human pose and shape from a
  single image.
\newblock In {\em ECCV}, 2016.

\bibitem{cao2016realtime}
Z.~Cao, T.~Simon, S.-E. Wei, and Y.~Sheikh.
\newblock Realtime multi-person 2d pose estimation using part affinity fields.
\newblock {\em CVPR}, 2017.

\bibitem{chen20163d}
C.-H. Chen and D.~Ramanan.
\newblock 3d human pose estimation= 2d pose estimation+ matching.
\newblock {\em CVPR}, 2017.

\bibitem{chen2014articulated}
X.~Chen and A.~L. Yuille.
\newblock Articulated pose estimation by a graphical model with image dependent
  pairwise relations.
\newblock In {\em NIPS}, 2014.

\bibitem{chen2017adversarial2}
Y.~Chen, C.~Shen, X.-S. Wei, L.~Liu, and J.~Yang.
\newblock Adversarial learning of structure-aware fully convolutional networks
  for landmark localization.
\newblock {\em arXiv preprint arXiv:1711.00253}, 2017.

\bibitem{chen2017adversarial}
Y.~Chen, C.~Shen, X.-S. Wei, L.~Liu, and J.~Yang.
\newblock Adversarial posenet: A structure-aware convolutional network for
  human pose estimation.
\newblock {\em ICCV}, 2017.

\bibitem{chu2016structured}
X.~Chu, W.~Ouyang, H.~Li, and X.~Wang.
\newblock Structured feature learning for pose estimation.
\newblock In {\em CVPR}, 2016.

\bibitem{chu2017multi}
X.~Chu, W.~Yang, W.~Ouyang, C.~Ma, A.~L. Yuille, and X.~Wang.
\newblock Multi-context attention for human pose estimation.
\newblock {\em CVPR}, 2017.

\bibitem{denton2015deep}
E.~L. Denton, S.~Chintala, and R.~Fergus.
\newblock Deep generative image models using a laplacian pyramid of adversarial
  networks.
\newblock In {\em NIPS}, 2015.

\bibitem{du2016marker}
Y.~Du, Y.~Wong, Y.~Liu, F.~Han, Y.~Gui, Z.~Wang, M.~Kankanhalli, and W.~Geng.
\newblock Marker-less 3d human motion capture with monocular image sequence and
  height-maps.
\newblock In {\em ECCV}, 2016.

\bibitem{fang2017learning}
H.~Fang, Y.~Xu, W.~Wang, X.~Liu, and S.-C. Zhu.
\newblock Learning knowledge-guided pose grammar machine for 3d human pose
  estimation.
\newblock {\em AAAI}, 2018.

\bibitem{ferrari20092d}
V.~Ferrari, M.~Mar{\'\i}n-Jim{\'e}nez, and A.~Zisserman.
\newblock 2d human pose estimation in tv shows.
\newblock {\em Statistical and Geometrical Approaches to Visual Motion
  Analysis}, 2009.

\bibitem{fischler1973representation}
M.~A. Fischler and R.~A. Elschlager.
\newblock The representation and matching of pictorial structures.
\newblock {\em IEEE Transactions on computers}, 1973.

\bibitem{ganin2015unsupervised}
Y.~Ganin and V.~Lempitsky.
\newblock Unsupervised domain adaptation by backpropagation.
\newblock In {\em ICML}, 2015.

\bibitem{goodfellow2014generative}
I.~Goodfellow, J.~Pouget-Abadie, M.~Mirza, B.~Xu, D.~Warde-Farley, S.~Ozair,
  A.~Courville, and Y.~Bengio.
\newblock Generative adversarial nets.
\newblock In {\em NIPS}, 2014.

\bibitem{hoffman2017cycada}
J.~Hoffman, E.~Tzeng, T.~Park, J.-Y. Zhu, P.~Isola, K.~Saenko, A.~A. Efros, and
  T.~Darrell.
\newblock Cycada: Cycle-consistent adversarial domain adaptation.
\newblock {\em arXiv preprint arXiv:1711.03213}, 2017.

\bibitem{huang2017stacked}
X.~Huang, Y.~Li, O.~Poursaeed, J.~Hopcroft, and S.~Belongie.
\newblock Stacked generative adversarial networks.
\newblock In {\em CVPR}, 2017.

\bibitem{ionescu2014human36m}
C.~Ionescu, D.~Papava, V.~Olaru, and C.~Sminchisescu.
\newblock Human3. 6m: Large scale datasets and predictive methods for 3d human
  sensing in natural environments.
\newblock {\em IEEE transactions on pattern analysis and machine intelligence},
  2014.

\bibitem{isola2017image}
P.~Isola, J.-Y. Zhu, T.~Zhou, and A.~A. Efros.
\newblock Image-to-image translation with conditional adversarial networks.
\newblock 2017.

\bibitem{karras2017progressive}
T.~Karras, T.~Aila, S.~Laine, and J.~Lehtinen.
\newblock Progressive growing of gans for improved quality, stability, and
  variation.
\newblock {\em arXiv preprint arXiv:1710.10196}, 2017.

\bibitem{li20143d}
S.~Li and A.~B. Chan.
\newblock 3d human pose estimation from monocular images with deep
  convolutional neural network.
\newblock In {\em ACCV}, 2014.

\bibitem{liang2017recurrent}
X.~Liang, Z.~Hu, H.~Zhang, C.~Gan, and E.~P. Xing.
\newblock Recurrent topic-transition gan for visual paragraph generation.
\newblock 2017.

\bibitem{liu2016coupled}
M.-Y. Liu and O.~Tuzel.
\newblock Coupled generative adversarial networks.
\newblock In {\em NIPS}, 2016.

\bibitem{martinez2017simple}
J.~Martinez, R.~Hossain, J.~Romero, and J.~J. Little.
\newblock A simple yet effective baseline for 3d human pose estimation.
\newblock {\em ICCV}, 2017.

\bibitem{mehta2017monocular}
D.~Mehta, H.~Rhodin, D.~Casas, P.~Fua, O.~Sotnychenko, W.~Xu, and C.~Theobalt.
\newblock Monocular 3d human pose estimation in the wild using improved cnn
  supervision.
\newblock In {\em 3D Vision (3DV)}, 2017.

\bibitem{mehta2017vnect}
D.~Mehta, S.~Sridhar, O.~Sotnychenko, H.~Rhodin, M.~Shafiei, H.-P. Seidel,
  W.~Xu, D.~Casas, and C.~Theobalt.
\newblock Vnect: Real-time 3d human pose estimation with a single rgb camera.
\newblock {\em ACM Transactions on Graphics}, 2017.

\bibitem{moreno20163d}
F.~Moreno-Noguer.
\newblock 3d human pose estimation from a single image via distance matrix
  regression.
\newblock {\em CVPR}, 2017.

\bibitem{newell2016stacked}
A.~Newell, K.~Yang, and J.~Deng.
\newblock Stacked hourglass networks for human pose estimation.
\newblock In {\em ECCV}, 2016.

\bibitem{nie2017monocular}
B.~X. Nie, P.~Wei, and S.-C. Zhu.
\newblock Monocular 3d human pose estimation by predicting depth on joints.
\newblock In {\em ICCV}, 2017.

\bibitem{pavlakos2016coarse}
G.~Pavlakos, X.~Zhou, K.~G. Derpanis, and K.~Daniilidis.
\newblock Coarse-to-fine volumetric prediction for single-image 3d human pose.
\newblock {\em CVPR}, 2017.

\bibitem{pishchulin2013poselet}
L.~Pishchulin, M.~Andriluka, P.~Gehler, and B.~Schiele.
\newblock Poselet conditioned pictorial structures.
\newblock In {\em CVPR}, 2013.

\bibitem{radford2015unsupervised}
A.~Radford, L.~Metz, and S.~Chintala.
\newblock Unsupervised representation learning with deep convolutional
  generative adversarial networks.
\newblock {\em ICLR}, 2016.

\bibitem{ramakrishna2012reconstructing}
V.~Ramakrishna, T.~Kanade, and Y.~Sheikh.
\newblock Reconstructing 3d human pose from 2d image landmarks.
\newblock {\em ECCV}, 2012.

\bibitem{ren2005recovering}
X.~Ren, A.~C. Berg, and J.~Malik.
\newblock Recovering human body configurations using pairwise constraints
  between parts.
\newblock In {\em ICCV}, 2005.

\bibitem{Sohn17}
K.~Sohn, S.~Liu, G.~Zhong, X.~Yu, M.-H. Yang, and M.~Chandraker.
\newblock Unsupervised domain adaption for face recognition in unlabeled
  videos.
\newblock In {\em ICCV}, 2017.

\bibitem{tekin2016structured}
B.~Tekin, I.~Katircioglu, M.~Salzmann, V.~Lepetit, and P.~Fua.
\newblock Structured prediction of 3d human pose with deep neural networks.
\newblock {\em BMVC}, 2016.

\bibitem{tekin2016direct}
B.~Tekin, A.~Rozantsev, V.~Lepetit, and P.~Fua.
\newblock Direct prediction of 3d body poses from motion compensated sequences.
\newblock In {\em CVPR}, 2016.

\bibitem{tian2010fast}
T.-P. Tian and S.~Sclaroff.
\newblock Fast globally optimal 2d human detection with loopy graph models.
\newblock In {\em CVPR}, 2010.

\bibitem{tome2017lifting}
D.~Tome, C.~Russell, and L.~Agapito.
\newblock Lifting from the deep: Convolutional 3d pose estimation from a single
  image.
\newblock {\em CVPR}, 2017.

\bibitem{tompson2015efficient}
J.~Tompson, R.~Goroshin, A.~Jain, Y.~LeCun, and C.~Bregler.
\newblock Efficient object localization using convolutional networks.
\newblock In {\em CVPR}, 2015.

\bibitem{toshev2014deeppose}
A.~Toshev and C.~Szegedy.
\newblock Deeppose: Human pose estimation via deep neural networks.
\newblock In {\em CVPR}, 2014.

\bibitem{tzeng2015simultaneous}
E.~Tzeng, J.~Hoffman, T.~Darrell, and K.~Saenko.
\newblock Simultaneous deep transfer across domains and tasks.
\newblock In {\em ICCV}, 2015.

\bibitem{tzeng2017adversarial}
E.~Tzeng, J.~Hoffman, K.~Saenko, and T.~Darrell.
\newblock Adversarial discriminative domain adaptation.
\newblock {\em CVPR}, 2017.

\bibitem{villegas2017learning}
R.~Villegas, J.~Yang, Y.~Zou, S.~Sohn, X.~Lin, and H.~Lee.
\newblock Learning to generate long-term future via hierarchical prediction.
\newblock {\em arXiv preprint arXiv:1704.05831}, 2017.

\bibitem{vondrick2016generating}
C.~Vondrick, H.~Pirsiavash, and A.~Torralba.
\newblock Generating videos with scene dynamics.
\newblock In {\em NIPS}, 2016.

\bibitem{WangECCV2016}
X.~Wang and A.~Gupta.
\newblock Generative image modeling using style and structure adversarial
  networks.
\newblock {\em ECCV}, 2016.

\bibitem{wang2017fast}
X.~Wang, A.~Shrivastava, and A.~Gupta.
\newblock A-fast-rcnn: Hard positive generation via adversary for object
  detection.
\newblock {\em CVPR}, 2017.

\bibitem{wei2016convolutional}
S.-E. Wei, V.~Ramakrishna, T.~Kanade, and Y.~Sheikh.
\newblock Convolutional pose machines.
\newblock In {\em CVPR}, 2016.

\bibitem{wei2017object}
Y.~Wei, J.~Feng, X.~Liang, M.-M. Cheng, Y.~Zhao, and S.~Yan.
\newblock Object region mining with adversarial erasing: A simple
  classification to semantic segmentation approach.
\newblock {\em CVPR}, 2017.

\bibitem{wu2016single}
J.~Wu, T.~Xue, J.~J. Lim, Y.~Tian, J.~B. Tenenbaum, A.~Torralba, and W.~T.
  Freeman.
\newblock Single image 3d interpreter network.
\newblock In {\em ECCV}, 2016.

\bibitem{yang2017learning}
W.~Yang, S.~Li, W.~Ouyang, H.~Li, and X.~Wang.
\newblock Learning feature pyramids for human pose estimation.
\newblock In {\em ICCV}, 2017.

\bibitem{yang2016end}
W.~Yang, W.~Ouyang, H.~Li, and X.~Wang.
\newblock End-to-end learning of deformable mixture of parts and deep
  convolutional neural networks for human pose estimation.
\newblock In {\em CVPR}, 2016.

\bibitem{yang2011articulated}
Y.~Yang and D.~Ramanan.
\newblock Articulated pose estimation with flexible mixtures-of-parts.
\newblock In {\em CVPR}, 2011.

\bibitem{zhang2016stackgan}
H.~Zhang, T.~Xu, H.~Li, S.~Zhang, X.~Huang, X.~Wang, and D.~Metaxas.
\newblock Stackgan: Text to photo-realistic image synthesis with stacked
  generative adversarial networks.
\newblock {\em ICCV}, 2017.

\bibitem{zhao2018pose}
M.~Zhao, T.~Li, M.~A. Alsheikh, Y.~Tian, H.~Zhao, D.~Katabi, and A.~Torralba.
\newblock Through-wall human pose estimation using radio signals.
\newblock In {\em CVPR}, 2018.

\bibitem{zhou2017towards}
X.~Zhou, Q.~Huang, X.~Sun, X.~Xue, and Y.~Wei.
\newblock Towards 3d human pose estimation in the wild: a weakly-supervised
  approach.
\newblock In {\em ICCV}, 2017.

\bibitem{zhou2016sparseness}
X.~Zhou, M.~Zhu, S.~Leonardos, K.~G. Derpanis, and K.~Daniilidis.
\newblock Sparseness meets deepness: 3d human pose estimation from monocular
  video.
\newblock In {\em CVPR}, 2016.

\bibitem{zhu2017unpaired}
J.-Y. Zhu, T.~Park, P.~Isola, and A.~A. Efros.
\newblock Unpaired image-to-image translation using cycle-consistent
  adversarial networks.
\newblock {\em ICCV}, 2017.

\end{thebibliography}
}

\end{document}